\documentclass{article}

     \PassOptionsToPackage{numbers, compress}{natbib}

     \usepackage[final]{neurips_2024}

\usepackage[utf8]{inputenc} 
\usepackage[T1]{fontenc}    
\usepackage{hyperref}       
\usepackage{url}            
\usepackage{booktabs}       
\usepackage{amsfonts}       
\usepackage{nicefrac}       
\usepackage{microtype}      
\usepackage{xcolor}         

\usepackage[ruled,vlined,linesnumbered]{algorithm2e}
\usepackage{graphicx} 
\usepackage{multirow}
\usepackage{array, hhline} 
\usepackage{booktabs} 
\newcolumntype{M}[1]{>{\centering\arraybackslash}m{#1}}
\usepackage{xcolor}

\title{Learning More Generalized Experts\\ 
by Merging Experts in Mixture-of-Experts}

\author{%
  Sejik Park \\
  Graduate School of AI, Korea Advanced Institute of Science and Technology (KAIST) \\
  Seoul, South Korea \\
  \texttt{sejik.park@kaist.ac.kr}
}

\begin{document}

\maketitle

\begin{abstract}
We observe that incorporating a shared layer in a mixture-of-experts can lead to performance degradation. This leads us to hypothesize that learning shared features poses challenges in deep learning, potentially caused by the same feature being learned as various different features. To address this issue, we track each expert's usage frequency and merge the two most frequently selected experts. We then update the least frequently selected expert using the combination of experts. This approach, combined with the subsequent learning of the router's expert selection, allows the model to determine if the most frequently selected experts have learned the same feature differently. If they have, the combined expert can be further trained to learn a more general feature. Consequently, our algorithm enhances transfer learning and mitigates catastrophic forgetting when applied to multi-domain task incremental learning.
\end{abstract}

\section{Introduction}
The Mixture-of-Experts (MoE) approach for class or task incremental learning has received increasing attention because using separate experts to solve different tasks enhances performance \citep{lin2024class, 2024divide}. However, in MoE, using or learning shared features is challenging due to the flexibility of deep learning \cite{park2024diverse}. Additionally, as shown in Table~\ref{tbl:intro}, increasing the shared parts between experts decreases the performance of the model \citep{2024divide}.

\begin{table} [h]
  \caption{\textbf{SEED \citep{2024divide} and DML \citep{zhang2018deep} with and without shared layers} This table shows the accuracy for the CIFAR100 dataset. It is divided into two sections, with the left side comparing the accuracy of increasing shared layers for the SEED, and the right side showing the accuracy of increasing shared layers for the DML algorithm or for a single model.}
  \label{tbl:intro}
  \centering
  \begin{tabular}{lcc||lccc}
    \hline
    Algorithm & 20-splits & 50-splits & Algorithm & Net1 & Net2 & Ensemble \\
    \hline
    \textbf{SEED \cite{2024divide}}  & \textbf{86.8} & \textbf{91.2} & \textbf{DML \cite{zhang2018deep}}                           & \textbf{70.97}  & \textbf{70.77}  & \textbf{72.65}\\
    SEED (1 shared)  & 86.7 & 91.2 & DML (shared) & 70.24  & 70.51  & 72.06\\
    SEED (11 shared) & 85.6 & 89.6 & Single model & 69.06  & -      & - \\
    SEED (21 shared) & 82.4 & 88.1 & \\
    \hline
  \end{tabular}
\end{table}

To elaborate on the results in Table~\ref{tbl:intro}, it can be observed that the method of selecting and training the optimal expert based on SEED's non-overlapping learning leads to a performance decline in class incremental learning on CIFAR100 as the number of shared layers (i.e., increasing the number of initial layers that are shared) increases across different class splits. Additionally, for the general classification task on CIFAR100, even in DML, which improves performance by ensembling two models and adding a distillation loss, there is a performance decline as more layers are shared (configured to not share only the last layer).

One reason for the difficulty of learning shared features could be the learning of duplicated features in various ways. For example, the model could learn the same features as distinct features because of the variance in the features. In other words, if the variance of a feature is large and the distribution of the training data is skewed, which could be seen as multiple distinct distributions, there is less opportunity for it to be learned as a single feature, and it may instead be learned as different features. For this reason, in the MoE, there could be various duplications at the feature level among different experts, consuming the capacity of the MoE to learn other information. Additionally, learning the same features differently complicates the model's integration with other features which could be helpful for stitching the information between the features.

\begin{center}
\begin{figure}[t]
    \begin{center}
    \includegraphics[width=0.9\linewidth]{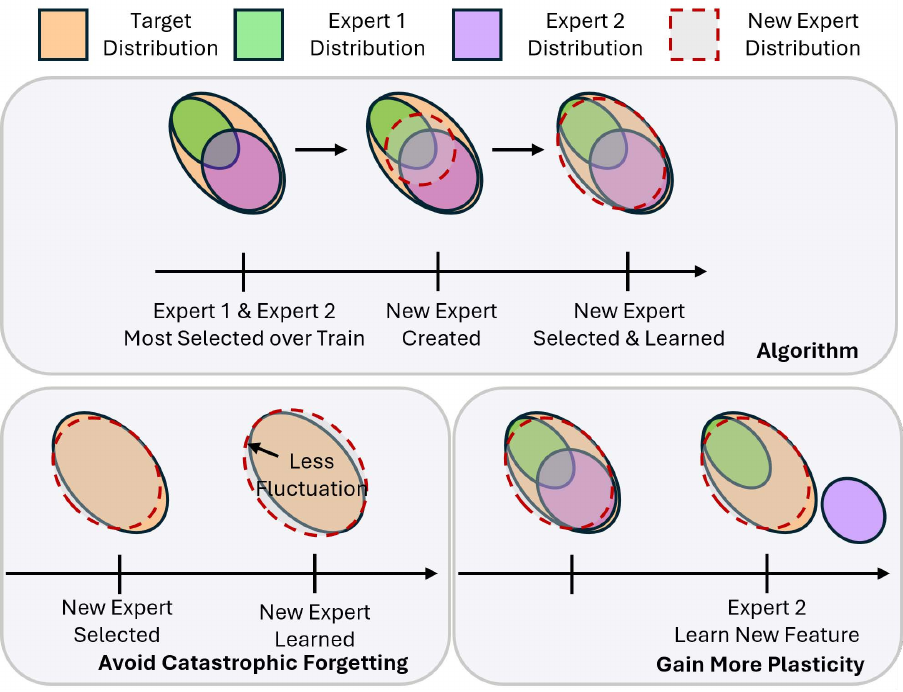}
    \end{center}
    \caption{\textbf{Overview of Principle} The top Algorithm block shows how MoE training proceeds in relation to creating an expert with a new distribution with our suggested learning algorithm. The bottom left Avoid Catastrophic Forgetting block explains why having a more general distribution of the new expert can help in consistently creating the same features. And the bottom right Gain More Plasticity block shows why the updates in our algorithm can be beneficial for additional learning.}
    \label{fig:introduction}
\end{figure}
\end{center}

With this purpose, as in the Figure~\ref{fig:introduction}, we propose an algorithm that merges different experts to create a new expert. In other words, this method incorporates newly created experts, formed by merging sufficiently trained experts, into the learning process to provide an opportunity to learn new distributions of features. Specifically, to select sufficiently trained experts, we utilize the frequently chosen experts from the router. Based on these selected experts, a new expert is created. To maintain the number of experts during training, the new expert updates the least frequently chosen expert. This approach allows the new expert to be selected by the router if it can learn a better distribution of features. In essence, this training algorithm adds a factor that enables more generalized learning of shared features which could reduce catastrophic forgetting and maintain plasticity.

To demonstrate the reduction of catastrophic forgetting and the maintenance of plasticity, we conducted experiments on multi-domain task incremental learning. In this setting, it necessitates efficient learning to remember features from past tasks and learn new features because multi-domain task incremental learning involves task changes, causing shifts in data distribution. With this approach, we show improvements in terms of accuracy related to transfer learning, average performance, and last performance. In other words, we indirectly demonstrates the advantages of our algorithm in reducing catastrophic forgetting and maintaining plasticity.

\section{Related Work}
Our method enhances the performance of MoE in multi-domain task incremental learning through weight merging. Therefore, in this related work section, we explain the related works in multi-domain task incremental learning and how MoE is applied in this setting. After that, we describe recent advancements in MoE and summarize related work concerning the merging of multiple model weights.

\textbf{Multi-domain Task Incremental Learning (MTIL)} studies the problems in the deep learning while the model learns from different dataset distribution, which could include new classes as the training processed \cite{zheng2023preventing}. In detail, 11 open-source datasets are used as sequential tasks. The order of the 11 datasets is Aircraft~\cite{maji2013fine}, Caltech101~\cite{fei2004learning}, CIFAR100~\cite{krizhevsky2009learning}, DTD~\cite{cimpoi2014describing}, EuroSAT~\cite{helber2019eurosat}, Flowers~\cite{nilsback2008automated}, Food~\cite{bossard2014food}, MNIST~\cite{deng2012mnist}, OxfordPet~\cite{parkhi2012cats}, Cars~\cite{krause20133d}, and SUN397~\cite{xiao2010sun}. Each task could have semantic overlap in terms of classes. The detailed description of each dataset is as follows.

\textbf{- Aircraft}~\cite{maji2013fine} contains 10,000 images of aircraft, with 100 model variants (e.g., Boeing 737-700). The train:validation:test split is 1:1:1. \\
\textbf{- Caltech101}~\cite{fei2004learning} includes objects belonging to 101 categories, with about 40 to 800 images per category (e.g., inline-skate). The total number of images is 8677, with a train:test split of 4:1. \\
\textbf{- CIFAR100}~\cite{krizhevsky2009learning} consists of 60,000 images of objects across 100 categories (e.g., beaver). The train:test split is 5:1. \\
\textbf{- DTD}~\cite{cimpoi2014describing} contains 5640 textural images divided into 47 categories (e.g., dotted). The train:validation:test split is 1:1:1. \\
\textbf{- EuroSAT}~\cite{helber2019eurosat} consists of 27,000 satellite images categorized into 10 classes (e.g., highway). The train:test split is 4:1. \\
\textbf{- Flowers}~\cite{nilsback2008automated} includes 7168 flower images divided into 102 categories (e.g., sunflower), with each category containing 40 to 258 images. The train:test split is approximately 1:6. \\
\textbf{- Food}~\cite{bossard2014food} comprises 101,000 images of food, categorized into 101 classes (e.g., pizza). The train:test split is 3:1. \\
\textbf{- MNIST}~\cite{deng2012mnist} consists of 70,000 digit images ranging from 0 to 9 (e.g., digit 1). The train:test split is 6:1. \\
\textbf{- OxfordPet}~\cite{parkhi2012cats} includes about 7400 images of pets across 37 breeds (e.g., Bengal). The train:test split is 1:1. \\
\textbf{- Cars}~\cite{krause20133d} is a dataset of 16,185 car images classified into 196 types (e.g., Acura Integra Type R 2001). The train:test split is approximately 1:1. \\
\textbf{- SUN397}~\cite{xiao2010sun} comprises 108,754 scene images categorized into 397 classes (e.g., outdoor). The train:test split is 4:1.

The main problems in MTIL are catastrophic forgetting and limited transfer ability. To address these issues, continual learning employs regularization methods, such as distillation loss over previous models \cite{li2017learning, ding2022don}, and re-learning past data using replay buffers \cite{rebuffi2017icarl}. Recent trends show efforts to leverage large pretrained models to benefit from their learned generalized features \cite{wortsman2022robust, zheng2023preventing, yu2024boosting}.

\textbf{Using Parts of Models for Incremental Learning} is a commonly utilized method because it preserves the learned parts to prevent forgetting \cite{zhu2022self}. For instance, a new task can be learned by creating a new part and setting up connections to prevent it from affecting the past model \cite{hu2023dense}. Alternatively, there are methods that ensure new parts of the model learn harmoniously with the existing parts \cite{zhou2024expandable, wang2022foster}. Additionally, algorithms such as mask learning or expert selection can be employed to selectively use parts of the model \cite{serra2018overcoming, lin2024class, 2024divide}.

Recently, to leverage the benefits of pretrained models, adapters with MoE algorithm have been proposed \cite{yu2024boosting}. This approach prevents catastrophic forgetting by preventing further training of task-wise frequently selected experts. In other words, utilizing the learning characteristics of MoE has been shown to be effective in preventing catastrophic forgetting in incremental learning. Then, the selection of experts is one of the most crucial factors that determine the characteristics of the MoE.

\textbf{Experts Selection in MoE} such as balancing selection \cite{kim2021scalable, roller2021hash, zhou2022mixture, mustafa2022multimodal}, task-specific selection \cite{li2022branch}, and learning-based methods \cite{shazeer2017, puigcerver2024from, qin2021bns, bengio2015conditional, rosenbaum2018routing, rosenbaum2019routing}, have been proposed. These methods consider aspects that can maintain plasticity or prevent catastrophic forgetting. However, these selection methods still have limitations in terms of stability during training.

To enhance the stability of training, a method to make hard selection softer when selecting experts has been introduced \cite{puigcerver2024from}. Another way to improve the training stability, multi-head MoE is also suggested to boost the activation of experts \cite{wu2024multi}. These studies aim to increase training stability based on the process of selecting experts, there is still limited research focusing on relationships among experts such as repetition. Our research has been conducted in relation to this issue by merging model weights among experts' weights.

\textbf{Merging Model Weight} has been extensively studied, ranging from leveraging the weights of various fine-tuned models \cite{choshen2022fusing, wortsman2022model} to using weights from the training process \cite{izmailov2018averaging}. This approach includes utilizing the weights of pre-trained models to generally enhance the performance of fine-tuned models \cite{wortsman2022robust, matena2022merging, ilharco2022patching}. Additionally, there is a method of using various weights by generating several different gradients with the addition of noise to the gradient \cite{von2020neural}.

For the MoE, a method utilizing the average of all experts has been researched \cite{huang2023experts}. While numerous methods have been explored to combine model weights to obtain more generalized weights, there is limited research that considers the possible aspect of learning repetitive experts in MoE. Therefore, we propose a method of merging weights of experts to generate a more generalized expert that provides an opportunity to learn a single expert that replaces experts learning the same meaning differently.

\section{Method}
\begin{center}
\begin{figure}[t]
    \begin{center}
    \includegraphics[width=0.9\linewidth]{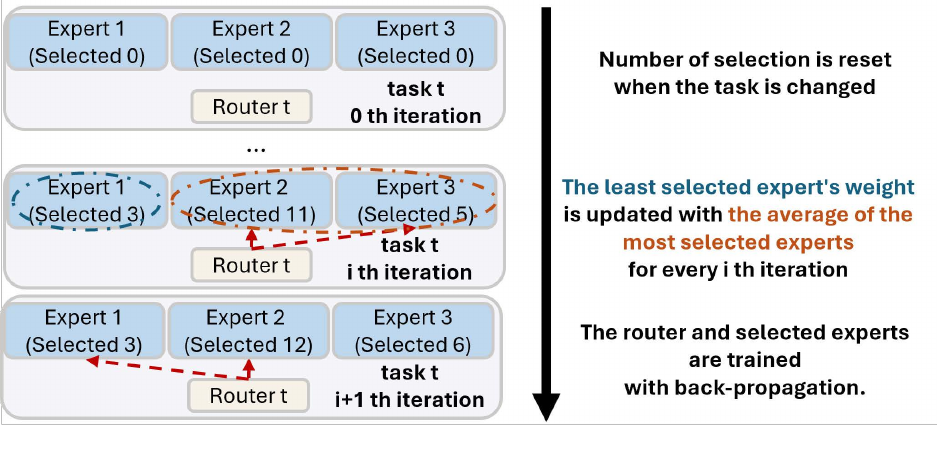}
    \end{center}
    \caption{\textbf{Overall Algorithm} Explain how our algorithm operates on an MoE with routing done by the top-k selection method when the number of experts is set to three.} 
    \label{fig:algorithm}
\end{figure}
\end{center}

We propose an algorithm that updates the least selected expert by averaging the weights of multiple experts as in Figure~\ref{fig:algorithm}. To explain in detail, in the MoE, the router selects experts using a top-K selection method. The number of selections made for each task by the top-K method is cumulatively stored. Periodically, based on the number of times each expert has been selected, the weights of the two most frequently chosen experts are averaged, and this average is used to replace the weight of the least selected expert.

We applied our algorithm to MA~\cite{yu2024boosting} which applies a Mixture-of-Experts (MoE) using lightweight adapters (LoRA~\cite{hu2021lora}) based on a pretrained transformer-based CLIP~\cite{dosovitskiy2020image}. For Multi-domain Task Incremental Learning (MTIL), MA prevents training on tasks that follow those where experts were most frequently selected via topK. The detail of our algorithm applied to MA for MTIL is shown in as Algorithm~\ref{alg:adapt_MA}. 

It comprises a number of tasks $T$ which are conducted in a predetermined order across various datasets. Each $t$-th task has a category set $C^t$. Each task $T^t$ consists of training data made up of multiple sets of an image $I$ and a class $y$. Additionally, since it utilizes CLIP, it employs an image encoder $F_I$ and a text encoder $F_T$. These encoders use a transfer encoder structure that includes several blocks, where each block contains multi-head attention with layer norm, MLP with layer norm, experts ${\epsilon_k}^{N_E}_{k=1}$, and routers $\mathrm{R}^t, t \in [1,T]$.

In the case of MoE, it uses the same input as the MLP with layer norm within the block, and the results are combined with the residual connection and the output of the MLP with layer norm. The detailed process of MoE involves selecting the top k logits for the experts through the router $R^t$ corresponding to the current task $t$, applying Softmax only to these logits, and using this as a kind of gate function to perform a weighted sum for the experts. This process is carried out for each block, and the experts and router in the MoE part, which is the adapter, are trained through back-propagation of the CLIP loss by performing contrastive learning on the final output.

\begin{algorithm}[t]
\label{alg:adapt_MA}
    \DontPrintSemicolon
	\caption{Our algorithm Adapted for MA Training}
        \textbf{Inputs:} Tasks $\{\mathrm{T^t}\}^T_{t=1}, \mathrm{T}^t = \{D^t = \{$image $I^t_i$, class $y^t_i\}^{N^t}_{i=1}$, Category set $C^t = \{c^t_j\}^{M^t}_{j=1} \}$ \\
        \textbf{Inputs:} Image encoder $F_I$, Text encoder $F_T$ (Each encoder includes several blocks that each block multi-head attention with layer norm, MLP with layer norm, experts $\{\epsilon_k\}^{N_E}_{k=1}$, and routers $\mathrm{R}^t, t \in [1,T]$) \\
        \textbf{Inputs:} Batch size $B$, Loss function $L$, Experts merging cycle $M$ \\

        \For{$t=1,2,...,T$}{
            \textcolor{red}{Reset the count of selection over each experts} \\
            \For{$i=1,2,...,(N^t//B)$}{
                \For{$b=1,2,...,B$}{
                    $s_b =  F_I(I^t_{(i-1)*B+b}) \cdot F_T(y^t_{(i-1)*B+b})$ \\
                    (Details of the block forward in $F_I$ and $F_T$ with input $x$ \\
                        $\qquad x^\prime = $ multi-head attention with layer norm $(x)$ \\
                        $\qquad x = x + x^\prime$ \\
                        $\qquad x^{\prime\prime} =$ MLP with layer norm $(x)$ \\
                        $\qquad W^t = Softmax(Topk(R^t(x[0])))$ \\
                        \textcolor{red}{$\qquad$ Update the count of selection over each experts} \\
                        $\qquad x^{\prime\prime\prime} = \sum^{N_E}_{k=1}W^t_k\epsilon_k(x)$ \\
                        $\qquad x = x + x^{\prime\prime} + x^{\prime\prime\prime}$ \\
                    )
                } 
                Back-propagation with $L(s_1, s_2, ..., s_B)$ \\
                \textcolor{red}{\If{$(i \% M) == 0$}{
      	        $t1, t2 = Top2($the count of selection over each experts$)$ \\
                    $b1 = Top1($the count of selection over each experts$)$ \\
                    $\epsilon_{b1} = (\epsilon_{t1} + \epsilon_{t2}) / 2$
                }}
            }
            $t_1, t_2, ..., t_k = Topk($the count of selection over each experts$)$ \\
            Frozen $\epsilon_{t_1}, \epsilon_{t_2}, ..., \epsilon_k$ for following tasks
        }
\end{algorithm}

In Algorithm~\ref{alg:adapt_MA}, parts related to our algorithm can be seen in red. This consists of measuring the number of times experts are trained by selected from the router per task and creating new expert weights based on these counts. The method of measuring the number of times an expert is trained involves resetting the count of selections for each expert when the task changes, and cumulatively measuring the number of selections for each iteration of the task. Then, the method of creating a new expert involves averaging the weights of the two most frequently selected experts at certain iterations, and the new expert replaces the weight of the expert with the least frequency of selection.

The main reason for selecting the most chosen experts is that these experts have been included in the selection to have the opportunity to learn features. Furthermore, the reason for updating the least chosen expert is to increase the utilization of experts that are less used in the model, thereby efficiently utilizing the overall model capacity. It also aims to minimize the impact on model updates since averaging the weights of the most selected experts may not always fall into significant areas.

Inference is conducted in the same manner as MA~\cite{yu2024boosting}. Specifically, each task is learned using an autoencoder, and the loss for each autoencoder is measured for the data to be classified. The router is then selected based on the task of the autoencoder with the smallest loss. Exceptionally, if the smallest loss measured by the autoencoder exceeds a certain threshold, it is considered out-of-distribution for the learned task, and the original CLIP is used.

\section{Experiments}
Our experiment was conducted on Multi-domain Task Incremental Learning (MTIL~\cite{zheng2023preventing}). For the all tasks in MTIL, each image is used after random resize crop and normalization. Then, for each task, training is conducted with random sampled data in batches of 64 for 1000 iterations. And the model's hyperparameters were set as follows: the number of experts, $N_E$, was set to 55. The purpose of providing a sufficient number of experts was to enable a clear distinction between meaningful and meaningless experts as the 11 tasks within MTIL were being learned. For the experts merging cycle $M$, we set it to 100 iterations. This is to ensure that the updated experts have sufficient time to learn.

Each training session utilized a GeForce 3090 and AMD EPYC 7402 24-Core Processor. And our method utilizes 22,785 MB of GPU memory and includes 486.9 MB of trainable parameters. Each iteration takes 1.55 seconds. Compared to MA~\cite{yu2024boosting}, which takes 1.48 seconds per iteration, our method requires an additional 0.07 seconds due to the extra algorithmic steps involved.

All experiments were conducted based on the publicly available GitHub repository of MA~\cite{yu2024boosting} (using same settings and hyperparameters which includes AdamW optimizer, label smoothing technique, threshold of out-of-distribution), with the number of experts modified to 55. Due to the large size of the model relative to the available number of GPUs, it was challenging to conduct experiments with multiple seeds. Therefore, the experiments were conducted with a single seed, and as a result, error margins were not included in the results.

For the qualitative results, transfer accuracy, average accuracy, and last accuracy are used for the metrics. Transfer accuracy refers to measuring the accuracy using a model trained up to the previous tasks, and it can demonstrate zero-shot transfer capability. Average accuracy means calculating the average of the accuracy measured for all tasks using the model trained up to each respective task. Last accuracy shows the accuracy for each task when the model has been trained up to the last task. Examining both average accuracy and last accuracy together allows for an indirect analysis of the degree of catastrophic forgetting.

\subsection{Results}
Table~\ref{tbl:detailed} compares the results between our proposed method and other related works applicable to MTIL. CLIP parts of Table serves as a kind of upper-bound, showing the zero-shot and fine-tune performance. Fine-tune includes training the entire model as well as training only the adapter. Then, our method has improved transfer accuracy, and average accuracy by 0.2\%, and 0.3\% respectively over the previous State-of-the-Art methods (MA~\cite{yu2024boosting}). This indicates that our model maintains plasticity and reduces the occurrence of catastrophic forgetting.

To analyze in more detail, we observe performance improvements in transfer accuracy of 0.3\% for DTD~\cite{cimpoi2014describing}, 1.2\% for Flowers~\cite{nilsback2008automated}, and 0.4\% for SUN397~\cite{xiao2010sun}. For the remaining datasets, the performance remains the same. This indicates that the transfer ability of the MoE has been improved. However, it is noticeable that the average accuracy and the last accuracy show different trends depending on the dataset. Therefore, the results in Table~\ref{tbl:detailed} might be due to increased variance caused by adding learning elements and performance improvements by chance due to the random seed setting.

\begin{table} [h]
  \caption{\textbf{Comparison with Various Methods on MTIL Benchmark} We label the best with bold about the average (exclude the top block which indicate the upper-bound by CLIP).}
  \label{tbl:detailed}
  \scriptsize
  \centering
  \begin{tabular}{llllllllllllll}
    \toprule
    & Method & \rotatebox{90}{Aircraft~\cite{maji2013fine}} & \rotatebox{90}{Caltech101~\cite{fei2004learning}} & \rotatebox{90}{CIFAR100~\cite{krizhevsky2009learning}} & \rotatebox{90}{DTD~\cite{cimpoi2014describing}} & \rotatebox{90}{EuroSAT~\cite{helber2019eurosat}} & \rotatebox{90}{Flowers~\cite{nilsback2008automated}} & \rotatebox{90}{Food~\cite{bossard2014food}} & \rotatebox{90}{MNIST~\cite{deng2012mnist}} & \rotatebox{90}{OxfordPet~\cite{parkhi2012cats}} & \rotatebox{90}{Cars~\cite{krause20133d}} & \rotatebox{90}{SUN397~\cite{xiao2010sun}} & Average \\
    \midrule
    \multirow{3}{*}{\rotatebox{90}{CLIP}}
& Zero-shot & 24.3 & 88.4 & 68.2 & 44.6 & 54.9 & 71.0 & 88.5 & 59.4 & 89.0 & 64.7 & 65.2 & 65.3 \\
& Full Fine-tune & 62.0 & 95.1 & 89.6 & 79.5 & 98.9 & 97.5 & 92.7 & 99.6 & 94.7 & 89.6 & 81.8 & 89.2 \\
& Fine-tune Adapter & 56.8 & 92.6 & 89.4 & 79.0 & 98.4 & 97.0 & 92.9 & 99.2 & 94.1 & 89.1 & 82.7 & 88.3 \\
    \midrule    
    \multirow{8}{*}{\rotatebox{90}{Transfer}}
& Continual-FT & & 67.1 & 46.0 & 32.1 & 35.6 & 35.0 & 57.7 & 44.1 & 60.8 & 20.5 & 46.6 & 44.6 \\
& LwF~\cite{li2017learning} & & 74.5 & 56.9 & 39.1 & 51.1 & 52.6 & 72.8 & 60.6 & 75.1 & 30.3 & 55.9 & 58.9 \\
& iCaRL~\cite{rebuffi2017icarl} & & 56.6 & 44.6 & 32.7 & 39.3 & 46.6 & 68.0 & 46.0 & 77.4 & 31.9 & 60.5 & 50.4 \\
& LwF-VR~\cite{ding2022don} & & 77.1 & 61.0 & 40.5 & 45.3 & 54.4 & 74.6 & 47.9 & 76.7 & 36.3 & 58.6 & 57.2 \\
& WiSE-FT~\cite{wortsman2022robust} & & 73.5 & 55.6 & 35.6 & 41.5 & 47.0 & 68.3 & 53.9 & 69.3 & 26.8 & 51.9 & 52.3 \\
& ZSCL~\cite{zheng2023preventing} & & 86.0 & 67.4 & 45.4 & 50.4 & 69.1 & 87.6 & 61.8 & 86.8 & 60.1 & 66.8 & 68.1 \\
& MA~\cite{yu2024boosting} & & 88.4&68.2&39.4&55.3&67.1&88.5&59.4&89.0&64.7&64.1&68.4 \\
& Ours & & 88.4&68.2&39.7&55.3&68.3&88.5&59.4&89.0&64.7&64.5&\textbf{68.6} \\
    \midrule
    \multirow{8}{*}{\rotatebox{90}{Average}}
& Continual-FT & 25.5 & 81.5 & 59.1 & 53.2 & 64.7 & 51.8 & 63.2 & 64.3 & 69.7 & 31.8 & 49.7 & 55.9 \\
& LwF~\cite{li2017learning} & 36.3 & 86.9 & 72.0 & 59.0 & 73.7 & 60.0 & 73.6 & 74.8 & 80.0 & 37.3 & 58.1 & 64.7 \\
& iCaRL~\cite{rebuffi2017icarl} & 35.5 & 89.2 & 72.2 & 60.6 & 68.8 & 70.0 & 78.2 & 62.3 & 81.8 & 41.2 & 62.5 & 65.7 \\
& LwF-VR~\cite{ding2022don} & 29.6 & 87.7 & 74.4 & 59.5 & 72.4 & 63.6 & 77.0 & 66.7 & 81.2 & 43.7 & 60.7 & 65.1 \\
& WiSE-FT~\cite{wortsman2022robust} & 26.7 & 86.5 & 64.3 & 57.1 & 65.7 & 58.7 & 71.1 & 70.5 & 75.8 & 36.9 & 54.6 & 60.7 \\
& ZSCL~\cite{zheng2023preventing} & 45.1 & 92.0 & 80.1 & 64.3 & 79.5 & 81.6 & 89.6 & 75.2 & 88.9 & 64.7 & 68.0 & 75.4 \\
& MA~\cite{yu2024boosting} & 49.6 & 93.1 & 83.7 & 67.9 & 80.6 & 82.6 & 88.7 & 73.6 & 89.1 & 68.5 & 65.5 & 76.6 \\
& Ours & 51.1 & 93.2 & 83.6 & 68.2 & 80.2 & 83.3 & 88.7 & 73.7 & 89.1 & 68.5 & 65.9 & \textbf{76.9}\\
    \midrule
    \multirow{8}{*}{\rotatebox{90}{Last}}
& Continual-FT & 31.0 & 89.3 & 65.8 & 67.3 & 88.9 & 71.1 & 85.6 & 99.6 & 92.9 & 77.3 & 81.1 & 77.3 \\
& LwF~\cite{li2017learning} & 26.3 & 87.5 & 71.9 & 66.6 & 79.9 & 66.9 & 83.8 & 99.6 & 92.1 & 66.1 & 80.4 & 74.6 \\ 
& iCaRL~\cite{rebuffi2017icarl} & 35.8 & 93.0 & 77.0 & 70.2 & 83.3 & 88.5 & 90.4 & 86.7 & 93.2 & 81.2 & 81.9 & 80.1 \\
& LwF-VR~\cite{ding2022don} & 20.5 & 89.8 & 72.3 & 67.6 & 85.5 & 73.8 & 85.7 & 99.6 & 93.1 & 73.3 & 80.9 & 76.6 \\
& WiSE-FT~\cite{wortsman2022robust} & 27.2 & 90.8 & 68.0 & 68.9 & 86.9 & 74.0 & 87.6 & 99.6 & 92.6 & 77.8 & 81.3 & 77.7 \\
& ZSCL~\cite{zheng2023preventing} & 40.6 & 92.2 & 81.3 & 70.5 & 94.8 & 90.5 & 91.9 & 98.7 & 93.9 & 85.3 & 80.2 & 83.6 \\
& MA~\cite{yu2024boosting} & 49.0 & 93.3 & 86.6 & 77.1 & 93.1 & 94.5 & 89.1 & 98.0 & 89.2 & 85.6 & 79.9 & \textbf{85.0} \\
& Ours & 50.7&92.7&86.5&77.0&92.6&94.1&89.0&98.5&89.2&85.5&79.8&\textbf{85.0} \\
    \bottomrule
  \end{tabular}
\end{table}

This becomes clearer when we look at the raw results in Table~\ref{tbl:raw_result}, which measures the performance of each task on the checkpoints learned for each task. Specifically, consistent or improved performance can only be confirmed for Aircraft~\cite{maji2013fine}, MNIST~\cite{deng2012mnist}, OxfordPet~\cite{parkhi2012cats}, and SUN397~\cite{xiao2010sun}. In other words, there is still a lack of consistent characteristics regarding performance improvements and declines.

\begin{table} [h]
  \caption{\textbf{Raw Experiments Results} It shows the accuracy on MTIL benchmark about MA~\cite{yu2024boosting} and our algorithm.}
  \label{tbl:raw_result}
  \small
  \centering
  \begin{tabular}{llllllllllllll}
    \toprule
    & Checkpoint & \rotatebox{90}{Aircraft} & \rotatebox{90}{Caltech101} & \rotatebox{90}{CIFAR100} & \rotatebox{90}{DTD} & \rotatebox{90}{EuroSAT} & \rotatebox{90}{Flowers} & \rotatebox{90}{Food} & \rotatebox{90}{MNIST} & \rotatebox{90}{OxfordPet} & \rotatebox{90}{Cars} & \rotatebox{90}{SUN397} \\
    \midrule
    \multirow{11}{*}{\rotatebox{90}{MA}}
& Aircraft & 51.3 & 88.4 & 68.2 & 44.7 & 55.3 & 71.0 & 88.5 & 59.5 & 89.0 & 64.7 & 65.2 \\
& Caltech101 & 51.1 & 94.6 & 68.2 & 35.2 & 55.3 & 69.7 & 88.5 & 59.5 & 89.0 & 64.7 & 62.7 \\
& CIFAR100 & 49.3 & 92.9 & 87.5 & 38.5 & 55.3 & 68.3 & 88.5 & 59.5 & 89.0 & 64.7 & 63.6 \\
& DTD & 49.1 & 93.3 & 87.2 & 79.9 & 55.3 & 63.2 & 88.5 & 59.5 & 89.0 & 64.7 & 64.1 \\
& EuroSAT & 49.1 & 93.3 & 87.3 & 80.2 & 95.5 & 63.4 & 88.5 & 59.5 & 89.0 & 64.7 & 64.1 \\
& Flowers & 49.4 & 94.1 & 87.2 & 79.6 & 96.4 & 97.5 & 88.4 & 59.5 & 89.0 & 64.7 & 64.2 \\
& Food & 49.5 & 93.8 & 87.3 & 78.9 & 95.6 & 96.9 & 89.1 & 59.5 & 89.0 & 64.7 & 64.3 \\
& MNIST & 49.6 & 93.5 & 87.1 & 77.5 & 95.8 & 95.8 & 89.1 & 98.4 & 89.0 & 64.7 & 64.3 \\
& OxfordPet & 49.1 & 93.4 & 86.8 & 78.0 & 95.0 & 94.6 & 89.0 & 98.2 & 89.2 & 64.7 & 64.2 \\
& StanfordCars & 48.6 & 93.0 & 86.8 & 77.8 & 94.0 & 94.2 & 89.0 & 98.3 & 89.2 & 85.9 & 64.3 \\
& SUN397 & 49.0 & 93.3 & 86.6 & 77.1 & 93.1 & 94.5 & 89.1 & 98.0 & 89.2 & 85.6 & 79.9 \\
     \midrule
    \multirow{11}{*}{\rotatebox{90}{Ours}}
& Aircraft & 52.3 & 88.4 & 68.2 & 44.7 & 55.3 & 71.0 & 88.5 & 59.5 & 89.0 & 64.7 & 65.2 \\
& Caltech101 & 51.3 & 94.0 & 68.2 & 35.7 & 55.3 & 68.4 & 88.5 & 59.5 & 89.0 & 64.7 & 63.5 \\
& CIFAR100 & 51.0 & 93.5 & 87.3 & 38.8 & 55.3 & 67.3 & 88.5 & 59.5 & 89.0 & 64.7 & 63.9 \\
& DTD & 50.6 & 94.1 & 87.2 & 80.0 & 55.3 & 67.3 & 88.5 & 59.5 & 89.0 & 64.7 & 64.4 \\
& EuroSAT & 51.0 & 94.0 & 87.3 & 79.8 & 94.8 & 67.5 & 88.5 & 59.5 & 89.0 & 64.7 & 64.5 \\
& Flowers & 51.1 & 94.3 & 87.2 & 80.1 & 95.2 & 97.7 & 88.3 & 59.5 & 89.0 & 64.7 & 64.6 \\
& Food & 50.7 & 94.1 & 87.1 & 78.6 & 94.3 & 97.0 & 89.1 & 59.5 & 89.0 & 64.7 & 64.8 \\
& MNIST & 51.3 & 93.4 & 87.0 & 78.5 & 94.8 & 96.3 & 89.0 & 98.7 & 89.0 & 64.7 & 64.8 \\
& OxfordPet & 51.2 & 93.1 & 86.8 & 78.5 & 94.9 & 95.4 & 89.0 & 98.5 & 89.2 & 64.7 & 64.8 \\
& StanfordCars & 50.8 & 93.5 & 86.8 & 78.1 & 94.5 & 94.4 & 89.0 & 98.6 & 89.2 & 85.5 & 64.8 \\
& SUN397 & 50.7 & 92.7 & 86.5 & 77.0 & 92.6 & 94.1 & 89.0 & 98.5 & 89.2 & 85.5 & 79.8 \\
    \bottomrule
  \end{tabular}
\end{table}

\begin{center}
\begin{figure}[t]
    \begin{center}
    \includegraphics[width=1.0\linewidth]{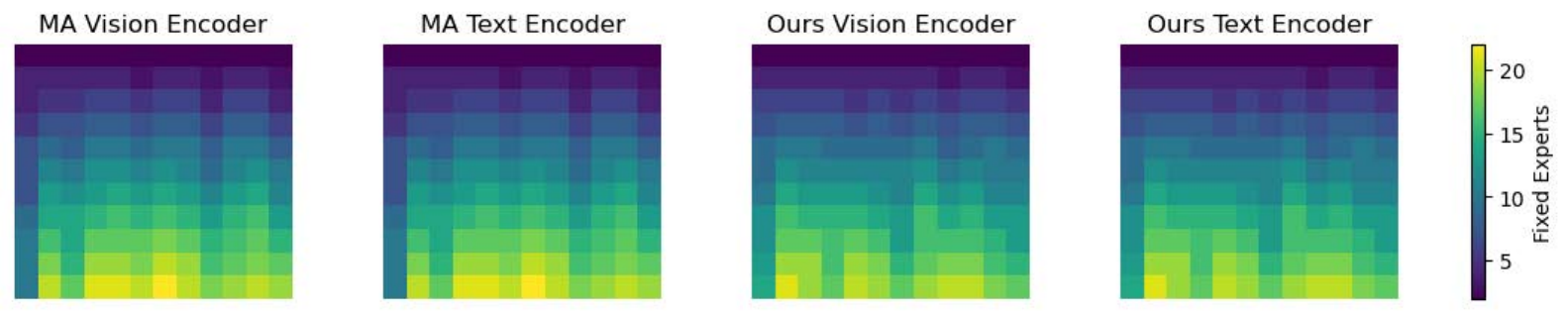}
    \end{center}
    \caption{\textbf{Number of Fixed Experts over Training} This figure visualizes the number of fixed experts in each block of the encoder (horizontal axis) as the task progresses (vertical axis).} 
    \label{fig:freeze}
\end{figure}
\end{center}

Furthermore, the trend of frozen experts can be observed in Figure~\ref{fig:freeze}. Since our expert freeze algorithm, which is applied through our method, progresses by selecting the most frequently chosen experts for each task, the overall lighter color indicates that our method requires relatively fewer experts. This suggests that it is able to better utilize the previously trained experts.

Another aspect that can be analyzed is the difference in the distribution of selected experts, specifically which level of blocks they are concentrated in. When looking at the relative outcomes for the final task, the conventional MA method often fixed the top-k selected experts in blocks positioned at intermediate stages. In contrast, our method shows a tendency to select more experts for the initial blocks. This suggests that our approach tends to create features differently for each task at the foundational level, which may help in generating consistent abstract features for diverse inputs in subsequent layers or lead to the usage of different features from the beginning.

Our results show that merging experts to create more generalized experts can improve qualitatively and quantitatively performance. However, there are aspects where performance improvements appear unstable. This instability may be due to the fact that the selection of experts to merge and the updating of experts are determined heuristically in our training algorithm. In other words, because there are few constraints on the experts being merged, it can be challenging to find advantages when creating a single expert based on two different experts. This could be why the experimental results showed only minor improvements. To address this, we could utilize the features shown in Figure~\ref{fig:freeze}, draw ideas from existing work related to merging weights, take inspiration from neural network diffusion in the brain, or leverage the similarity in expert selection over following MoE layers.

\section{Limitations} 
Our limitations can be largely divided into those related to the dataset and those related to the proposed algorithm. Regarding the dataset, because we hoped the algorithm would utilize features in a generalized manner, there needs to be some commonality between tasks to determine if transfer learning works well between them. However, currently, many of the tasks are composed of datasets that were specifically collected, which may pose a limitation. As for the algorithm, it is based on heuristic ideas, so direct theoretical or experimental analysis is needed to determine if the model can indeed demonstrate the advantages of this approach.

To explain the limitations related to the data in more detail: We utilized MTIL to undertake tasks that consider various distributional changes. However, there are limitations in the extent to which the distributions of tasks included in the utilized MTIL share commonalities. In this regard, larger datasets such as ImageNet can be used. For example, by creating tasks with differing distribution commonalities from a large dataset using MTIL and analyzing them, it would be possible to conduct an in-depth analysis of whether features were learned in a generalized manner, as well as the necessity and relevance of the data.

One of the limitations related to the algorithm is that the selection of experts to merge is based on simple heuristics. To make this decision using information from the model itself, one could utilize the experts selected consistently through back-propagation (i.e., the same experts used in subsequent layers). Additionally, it is necessary to analyze whether updating the less frequently selected experts has less impact on the overall training. Moreover, analyzing the impact of our proposed algorithm on MoE methods other than MA would allow us to understand whether the influence varies depending on the MoE configuration.

Another limitation is that due to the lack of GPU resources, we could not analyze whether the differences observed are statistically significant. Conducting experiments with various seeds to demonstrate statistically significant differences would be a meaningful direction for future research.

\section{Conclusion}
We propose a method to address the difficulty in learning shared features in MoE by merging experts to create a new expert. The difficulty in learning shared features appears to be indirectly reflected in the performance degradation as the number of shared layers increases. To address this, we included an expert which could have more generalized shared features by merging well-trained experts in the training process of MoE applied to multiple layers. This approach shows negligible computational differences but provides benefits in terms of transfer learning performance and average accuracy.

The reason why creating a new expert by merging experts and including it in the training can be beneficial is that the newly formed expert is shaped in a space that allows it to learn more generalized features than the experts used in the merging process. This helps in avoiding catastrophic forgetting and maintaining plasticity. Additionally, generating more generalized features provides advantages for feature stitching. However, the current method is heuristic and limited to specific moments, as it is based on the number of times the router selects experts for merging. This approach may have limitations in creating an expert that can generate more generalized features. To address these limitations, we plan to leverage the similarity in the utilization of experts in following MoE layers in future work, or apply methods to stabilize the data distribution in existing weight merging techniques.

Our experiments were conducted on Multi-domain Task Incremental Learning (MTIL). For the MTIL tasks, we utilized datasets readily accessible from torchvision. However, there was a drawback in that the data distribution differed significantly between tasks. This approach may be insufficient for analyzing factors such as transfer learning ability. To address this, we plan to further study the impact on learning when controlling the degree of feature variation between tasks by dividing large datasets like ImageNet based on label similarity, ensuring that the dataset composition reflects both high and low degrees of feature changes between tasks in MTIL.

Additionally, we aim to empirically demonstrate that merging experts can be generally applied to MoE through experiments with more diverse MoE configurations. This means we will analyze how our method is affected by the way experts are structured within the model, the number of experts, and factors external to MoE. We also plan to establish methods to efficiently analyze statistical significance.

If someone is interested in related research, would like to discuss this topic further, or are willing to support the research with GPUs, please contact us via the shared email. For requesting code or trained weights, please reach out through email as well. Any advice on improving our paper is also greatly appreciated.

\newpage
\bibliographystyle{plain}
\bibliography{reference}



\end{document}